\documentclass{bmvc2k}

\usepackage{amsmath,mathtools,amssymb}
\usepackage{hyperref}
\title{Virtual to Real Reinforcement Learning for Autonomous Driving}

\addauthor{Xinlei Pan$^*$}{xinleipan@berkeley.edu}{1}
\addauthor{Yurong You$^*$}{yurongyou@sjtu.edu.cn}{2}
\addauthor{Ziyan Wang}{zy-wang13@mails.tsinghua.edu.cn}{3}
\addauthor{Cewu Lu}{lu-cw@cs.sjtu.edu.cn}{2}

\addinstitution{
 University of California, Berkeley\\
 Berkeley, CA, USA
}
\addinstitution{
 Shanghai Jiao Tong University \\ 
 Shanghai, China
}
\addinstitution{
 Tsinghua University \\ 
 Beijing, China
}

\runninghead{Pan,You,Wang,Lu}{Virtual to Real Reinforcement Learning}

\begin{document}
\maketitle
\begin{abstract}
Reinforcement learning is considered as a promising 
direction for driving policy learning. However, 
training autonomous driving vehicle with reinforcement 
learning in real environment involves non-affordable 
trial-and-error. It is more desirable to first train 
in a virtual environment and then transfer to the real 
environment. In this paper, we propose a novel realistic 
translation network to make model trained in virtual 
environment be workable in real world. 
The proposed network can convert non-realistic 
virtual image input into a realistic one with similar 
scene structure. Given realistic frames as input, 
driving policy trained by reinforcement learning
 can nicely adapt to real world driving. Experiments 
 show that our  proposed virtual to real (VR) 
 reinforcement learning (RL) works pretty well. 
To our knowledge, this is the first successful case 
of driving policy trained by reinforcement learning 
that can adapt to \textbf{real world} driving data. 
\end{abstract}

Autonomous driving aims to make a vehicle sense its 
environment and navigate without human input. To 
achieve this goal, the most important task is to 
learn the driving policy that automatically outputs 
control signals for steering wheel, throttle, brake, 
etc., based on observed surroundings. The straight-forward
idea is end-to-end supervised learning 
\cite{nvidia,jianxiong}, which trains a neural network
model mapping visual input directly to action output, 
and the training data is labeled image-action pairs. 
However, supervised approach usually requires large
amount of data to train a model \cite{xu2016end} that
can generalize to different environments. Obtaining
such amount of data is time consuming and requires 
significant human involvement. By contrast, 
reinforcement learning learns by a trial-and-error fashion, 
and does not require explicit supervision from human. Recently,
reinforcement learning has been considered as a promising
technique to learn driving policy due to its expertise in 
action planing \cite{sallab2016end,
DBLP:journals/corr/Shalev-ShwartzS16a,lillicrap2015continuous}. 

\begin{figure*}[t]
\includegraphics[width=\linewidth]{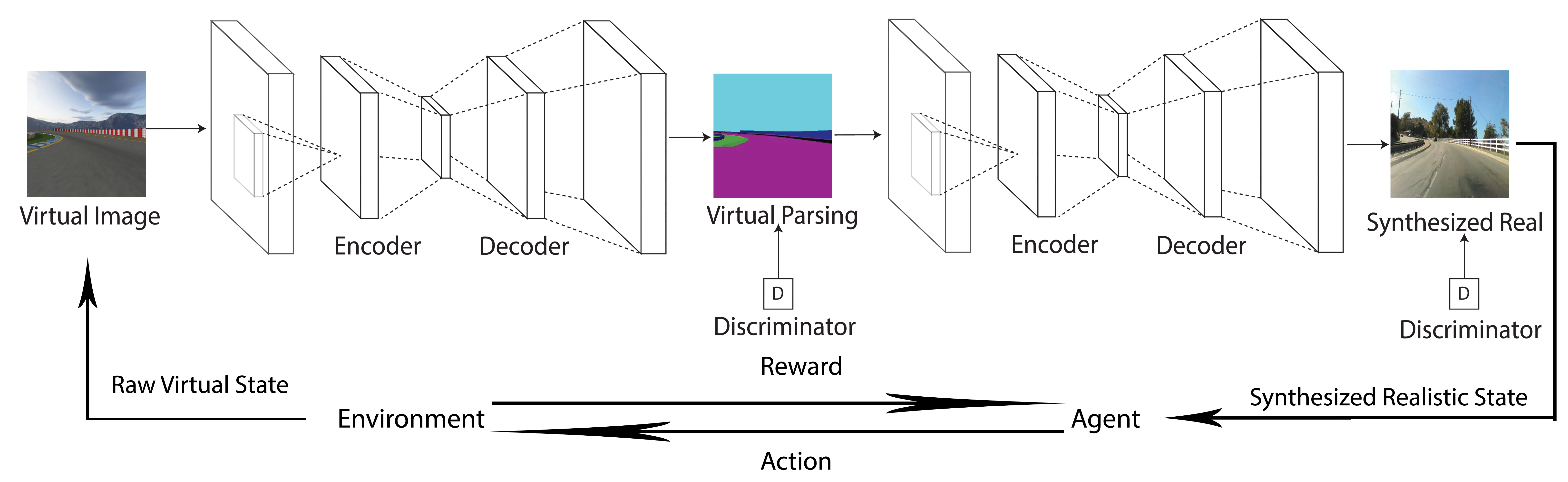}
\caption{Framework for virtual to real
reinforcement learning for autonomous driving.
Virtual images rendered by a simulator (environment)
are first segmented to scene parsing representation 
and then translated to synthetic realistic images by
the proposed image translation network (VISRI). Agent
observes synthetic realistic images and takes actions.
Environment will give reward to the agent. Since the
agent is trained using realistic images that are visually
similar to real world scenes, it can nicely adapt to real world
driving. }
\label{fig1}
\end{figure*}

However, reinforcement learning requires agents to interact
with environments, and undesirable driving actions would happen. 
Training autonomous driving cars in real world will 
cause damages to vehicles and the surroundings. Therefore,
most of current research in autonomous driving with 
reinforcement learning focus on simulations
\cite{A3C,DBLP:journals/corr/Shalev-ShwartzS16a,lillicrap2015continuous}
rather than training in real world. While an agent trained
with reinforcement learning achieves near human-level
driving performance in virtual world \cite{A3C}, it may
not be applicable to real world driving environment,
since the visual appearance of virtual simulation environment 
is different from that of real world driving scene.

While virtual driving scenes have a different visual 
appearance compared with real driving scenes, they
share similar scene parsing structure. For example,
virtual and real driving scenes may all have roads,
trees, buildings, etc., though the textures may be
significantly different. Therefore,
it is reasonable that by translating virtual images
to their realistic counterparts, we can obtain a simulation
environment that looks very similar to the real world 
in terms of both scene parsing structure and object
appearance. Recently, generative adversarial network (GAN) 
\cite{NIPS2014_5423}
has drawn a lot of attention in image generation.
The work by \cite{IsolaZZE16} proposed an image-to-image
translation network that can translate images from
one domain to another using paired data from both
domains. However, it is very hard to find paired 
virtual-real world images for driving, making 
it difficult to apply this method to our case 
of translating virtual driving images to realistic ones. 

In this paper, we propose a 
\emph{realistic translation network} to help train
self-driving car entirely in virtual world that can
adapt to real world driving environment.
Our proposed framework (shown in Figure ~\ref{fig1}) 
converts virtual images rendered by the simulator to 
a realistic one and train the reinforcement learning
agent with the synthesized realistic images. 
Though virtual and realistic images 
have a different visual appearance, they share a common 
scene parsing representation (segmentation map of 
roads, vehicles  etc.). Therefore, we can translate virtual images to 
realistic images by using scene parsing representation as 
the interim. This insight is similar to natural 
language translation, where semantic meaning is the 
interim between different languages. Specifically, 
our realistic translation network includes two modules. 
The first one is a virtual-to-parsing or virtual-to-segmentation 
module that produces a scene parsing representation of 
input virtual image. The second one is a parsing-to-real 
network that translates scene parsing representations 
into realistic images. With realistic translation network, 
reinforcement learning model learnt on the realistic 
driving data can nicely apply to real world driving.

To demonstrate the effectiveness of our method, 
we trained our reinforcement learning model by using 
the realistic translation network to filter virtual
images to synthetic realistic images and feed these realistic
images as state inputs. We further compared with
supervised learning and other 
reinforcement learning approaches that use domain
randomization \cite{DBLP:journals/corr/SadeghiL16}. 
Our experiments illustrate 
that a reinforcement learning model trained with 
translated realistic images has better 
performance than reinforcement learning model 
trained with only virtual input and virtual to real
reinforcement learning with domain randomization.

\section{Related Work}

 \textbf{Supervised Learning for Autonomous Driving}.
 Supervised learning methods are obviously 
 straightforward ways to train autonomous vehicles. 
 ALVINN \cite{pomerleau1989alvinn} provides an early 
 example of using neural network for autonomous 
 driving. Their model is simple and direct, which 
 maps image inputs to action predictions
 with a shallow network. Powered by deep learning 
 especially a convolutional neural network, NVIDIA 
 \cite{nvidia} recently provides an attempt to leverage
 driving video data for simple lane following task. 
 Another work by \cite{jianxiong} learns a mapping 
 between input images to a number  of key perception 
 indicators, which are closely related to the 
 affordance of a  driving state. However, the learned 
 affordance must be associated with actions  
 through hand-engineered rules. These supervised 
 methods work relatively  well in simple tasks such as 
 lane-following and driving on a highway. On the other 
 hand, imitation learning can also be regarded as 
 supervised learning approach \cite{zhang2016query}, 
 where the agent observes the demonstrations performed 
 by some experts and learns to imitate the actions of
 the experts. However, an intrinsic shortcoming of 
 imitation learning is that it has the 
 covariate shift problem \cite{ross2010efficient} and
 can not generalize very well to scenes never experienced
 before.

\textbf{Reinforcement Learning for Autonomous Driving}. 
Reinforcement learning has been applied to a wide variety of 
robotics related tasks,  such as computer games \cite{mnih2015human},  
robot locomotion \cite{kohl2004policy,endo2008learning}, 
and autonomous driving  \cite{abbeel2007application, 
DBLP:journals/corr/Shalev-ShwartzS16a}. One of the challenges 
 in practical real-world applications of reinforcement 
 learning is the high-dimensionality of state space as well as 
 the non-trivial large action range. Developing an optimal policy over 
 such high-complexity space is time consuming. Recent work 
 in deep reinforcement  learning has made great progress 
 in learning in a high dimensional space with the power of 
 deep neural networks \cite{koutnik2013evolving, mnih2015human, schulman2015trust,lillicrap2015continuous,A3C}. However, 
 both deep Q-learning method \cite{mnih2015human} and 
 policy gradient method \cite{lillicrap2015continuous} 
 require the agent to interact with the environment
 to get reward and feedback. However, it is unrealistic
 to train autonomous vehicle with reinforcement learning in a real
 world environment since the car may hurt its surroundings once
 it takes a wrong action. 

\textbf{Reinforcement Learning in the Wild}. Performing reinforcement 
learning with a car driving simulator and transferring learned models to 
the real environment could enable faster, lower-cost training, and it is much 
safer than training with a real car. However, real-world driving challenge usually 
spans a diverse range, and it is often significantly different 
from the training environment in a car driving simulator in terms 
of their visual appearance. Models trained purely on virtual data 
do not generalize well to real images \cite{christiano2016transfer, 
tzeng2016adapting}. Recent progress of transfer and domain adaptation 
learning in robotics provides examples of simulation-to-real 
reinforcement training \cite{DBLP:journals/corr/RusuVRHPH16,
gupta2017learning,tobin2017domain}. These models either first 
train a model in virtual environment and then fine-tune in the 
real environment \cite{DBLP:journals/corr/RusuVRHPH16}, or learn 
an alignment between virtual images and real images by finding 
representations that are shared between the two domains 
\cite{tzeng2016adapting}, or use randomized rendered virtual 
environments to train and then test in real environment 
\cite{DBLP:journals/corr/SadeghiL16,tobin2017domain}. The 
work of \cite{DBLP:journals/corr/RusuVRHPH16} proposes to use
progressive network to transfer network weights from model
trained on virtual data to the real environment and then 
fine-tune the model in a real setting. The training time 
in real environment has been greatly reduced by first 
training in a virtual environment. However, it is still 
necessary to train the agent in the real environment, 
thus it does not solve the critical problem of avoiding 
risky trial-and-error in real world. Methods that try 
to learn an alignment between virtual images and real 
images could fail to generalize to more complex scenarios, 
especially when it is hard to find a good alignment 
between virtual images and real images. As a more recent 
work, \cite{DBLP:journals/corr/SadeghiL16} proposed a new 
framework for training a reinforcement learning agent with 
only a virtual environment. Their work proved the possibility 
of performing collision-free flight in real world with training 
in 3D CAD model simulator. However, as mentioned in the 
conclusion of their paper \cite{DBLP:journals/corr/SadeghiL16}, 
the manual engineering work to design suitable training 
environments is nontrivial, and it is more reasonable to 
attain better results by combining simulated training with 
some real data, though it is unclear from their paper
how to combine real data with simulated training.

\textbf{Image Synthesis and Image Translation}. 
Image translation aims to predict image in some specific 
modality, given an image from another modality. 
This can be treated as a generic method as it predicts 
pixels from pixels. Recently, the community has made significant 
progress in generative approaches, mostly based on generative 
adversarial networks \cite{NIPS2014_5423}. To name a few, 
the work of \cite{3dgan} explored the use of VAE-GAN \cite{DBLP:journals/corr/LarsenSW15} 
in generating 3D voxel 
models, and the work of \cite{Wang_SSGAN2016} proposed a cascade GAN to 
generate natural images by structure and style. More recently, the work of \cite{IsolaZZE16} 
developed a general and simple 
framework for image-to-image translation which can handle 
various pixel level generative tasks like semantic 
segmentation, colorization, rendering edge maps, etc.

\textbf{Scene Parsing}. 
One part of our network is the semantic image
segmentation network. There are already many great 
works in the field of semantic image segmentation. 
Many of them are based on deep convolutional neural network or
fully convolutional neural network \cite{Long_2015_CVPR}. 
In this paper, we use the SegNet for image segmentation, 
the structure of the network is revealed in \cite{badrinarayanan2015segnet}, 
which is composed of two main parts. The first part is an encoder, 
which consists of Convolutional, Batch Normalization, ReLU 
and max pooling layers. The second part is a decoder, which 
replaces the pooling layers with upsampling layers.


\begin{figure*}[t]
\centering
\includegraphics[width=0.46\linewidth,height=0.23\linewidth]{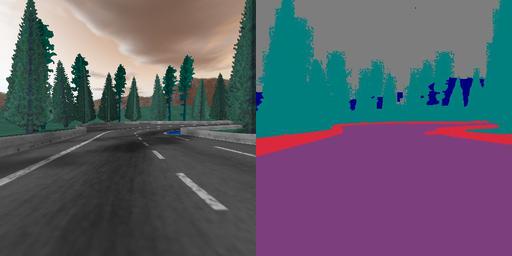}
\includegraphics[width=0.46\linewidth,height=0.23\linewidth]{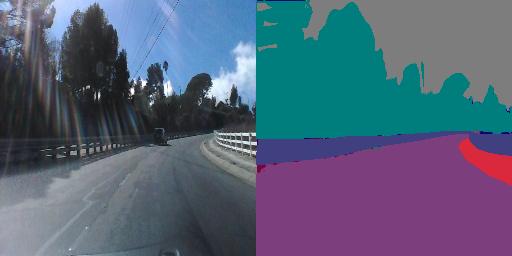}
\includegraphics[width=0.46\linewidth,height=0.23\linewidth]{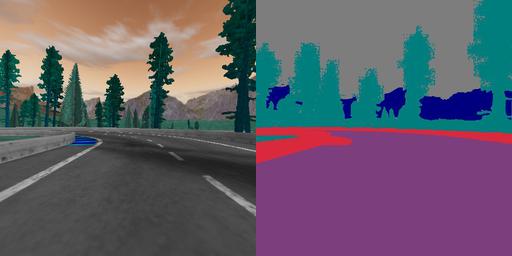}
\includegraphics[width=0.46\linewidth,height=0.23\linewidth]{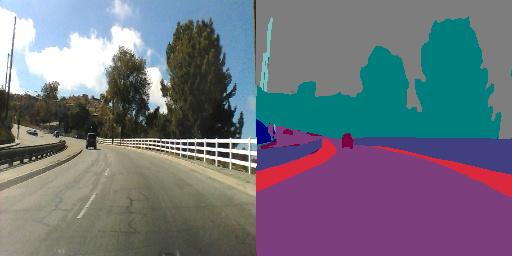}
\includegraphics[width=0.46\linewidth,height=0.23\linewidth]{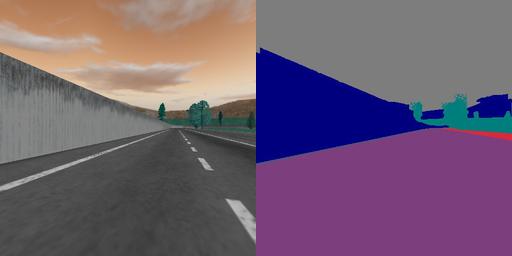}
\includegraphics[width=0.46\linewidth,height=0.23\linewidth]{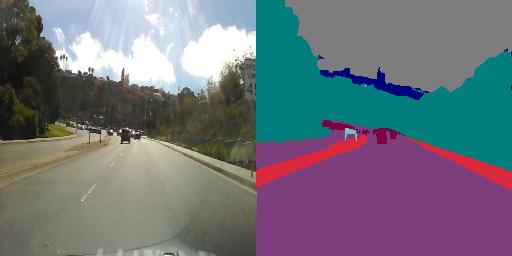}
\caption{Example image segmentation for both virtual world images (Left 1 and Left 2)
and real world images (Right 1 and Right 2). }
\label{fig2}
\end{figure*}

\section{Reinforcement Learning in the Wild}
We aim to successfully apply a driving model trained entirely 
in virtual environment to real-world driving challenges. 
One of the major gaps is that what 
the agent observes are frames rendered by a simulator, 
which are different from real world frames in terms of their 
appearance. Therefore, we proposed a \emph{realistic translation network}
to convert virtual frames to realistic ones. Inspired by the work 
of image-to-image translation network \cite{IsolaZZE16}, 
our network includes two modules, namely virtual-to-parsing 
and parsing-to-realistic network. The first one maps virtual 
frame to scene parsing image. The second one translates scene 
parsing to realistic frame with similar scene structure as 
the input virtual frame. These two modules generate
realistic frames that maintain the scene parsing 
structure of input virtual frames. The architecture of 
realistic translation network is illustrated on Figure
~\ref{fig1}. Finally, we train a self-driving agent 
using reinforcement learning method on 
realistic frames obtained by realistic translation network. The approach we 
adopt is developed by \cite{A3C}, where they use the asynchronous 
actor-critic reinforcement learning algorithm to train a self-driving vehicle 
in the car racing simulator TORCS \cite{wymann2000torcs}. 
In this section, we will first present proposed realistic 
translation network and then discuss how to train driving 
agent under a reinforcement learning framework.

\subsection{Realistic Translation Network}
As there is no paired virtual and real world image,
a direct mapping from virtual world image to
real world image using \cite{IsolaZZE16} would be awkward. 
However, as these two types of images both express 
driving scene, we can translate them by using scene parsing 
representation. Inspired by \cite{IsolaZZE16}, our realistic 
translation network is composed of two image translation networks, 
where the first image translation network translates virtual 
images to their segmentations, and the second image translation
network translates segmented images to their real world
counterparts. 

The image-to-image translation network proposed by
\cite{IsolaZZE16} is basically a conditional GAN.
The difference between traditional GANs and 
conditional GANs is that GANs learn a mapping 
from random noise vector $z$ to output image 
$s: G: z\rightarrow s$, while conditional 
GANs take in both an image
$x$ and a noise vector $z$, and generate another 
image $s: G:\{x,z\}\rightarrow s$, where $s$ is 
usually in a different domain 
compared with $x$ (For example, translate images
to their segmentations). 

The objective of a conditional GAN can be expressed as,
\begin{equation}
\begin{split}
\mathcal{L}_{cGAN}(G, D) = &\mathbb{E}_{x,s\sim p_{data}(x,s)}[\log D(x,s)] \\
& + \mathbb{E}_{x\sim p_{data}(x), z\sim p_z(z)}[\log (1 - D(x, G(x, z)))], \\
\end{split}
\end{equation}

where $G$ is the generator that tries to 
minimize this objective and $D$ is the 
adversarial discriminator that acts against
$G$ to maximize this objective. In other
words, $G^{*} = \arg\min_{G}\max_{D}\mathcal{L}
_{cGAN}(G,D)$. In order to suppress blurring,
a L1 loss regularization term is added, which 
can be expressed as,
\begin{equation}
\mathcal{L}_{L1}(G) = \mathbb{E}_{x, s\sim p_{data}(x,s), z\sim p_z(z)} 
[\| s - G(x, z)\|_1].
\end{equation}
Therefore, the overall objective for the image-to-image
translation network is,
\begin{equation}
G^{*} = \arg\min_{G}\max_{D}\mathcal{L}_{cGAN}(G, D)
+\lambda\mathcal{L}_{L1}(G),
\label{eqloss}
\end{equation}
where $\lambda$ is the weight of regularization.

Our network consists of two image-to-image
translation networks, both networks use the
same loss function as equation~\ref{eqloss}. The
first network translates virtual images $x$
to their segmentations $s: G_1: \{x,z_1\}\rightarrow s$,
and the second network translates segmented
images $s$ into their realistic counterparts
$y: G_2: \{s,z_2\}\rightarrow y$, where $z_1, z_2$
are noise terms to avoid deterministic outputs.
As for GAN neural network structures, we use the 
same generator and discriminator architectures
as used in \cite{IsolaZZE16}.

\subsection{Reinforcement Learning for Training 
a Self-Driving Vehicle}
We use a conventional RL solver Asynchronous Advantage Actor-Critic 
(A3C)\cite{A3C} to train the self driving vehicle, which has performed well
on various machine learning tasks. A3C algorithm is a fundamental Actor-Critic 
algorithm that combines several classic reinforcement learning algorithms
with the idea of asynchronous parallel threads. Multiple threads run at the 
same time with unrelated copies of the environment, generating their own 
sequences of training samples. Those actors-learners proceed as though 
they are exploring different parts of the unknown space. For one thread, 
parameters are synchronized before an iteration of learning and updated 
after finishing it. The details of A3C algorithm implementation can be found
in \cite{A3C}. 

In order to encourage the agent to drive faster and avoid collisions, 
we define the reward function as
\begin{align}
	r_t = \left\{
	\begin{array}{ll}
		(v_t \cdot \cos\alpha - \mathrm{dist}_{\text{center}}^{(t)})\cdot \beta & \text{no collision,}\\
		\gamma & \text{collision},
			\end{array}
			\right.
			\label{eq::reward}
		\end{align}
where $v_t$ is the speed (in $m/s$) of the agent at 
time step $t$, $\alpha$ is the angle (in rad) between 
the agent's speed and the tangent line of the track, 
and $\mathrm{dist}_{\text{center}}^{(t)}$ is the distance 
between the center of the agent and the middle of 
the track. $\beta, \gamma$ are constants and are determined
at the beginning of training. We take $\beta = 0.006, 
\gamma = -0.025$ in our training. 

\begin{figure*}[t]
\centering
\includegraphics[width=\linewidth]{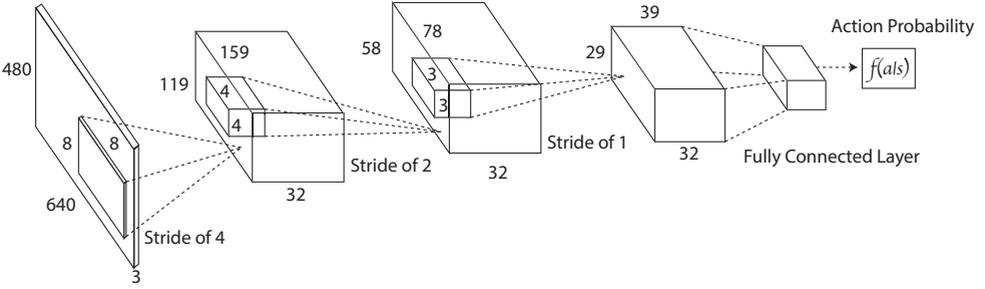}
\caption{Reinforcement learning network architecture. The network is an
end-to-end network mapping state representations to action probability
outputs.}
\label{netarch}
\end{figure*}

\begin{figure*}
\centering
\includegraphics[width=0.24\linewidth, height=0.125\linewidth]{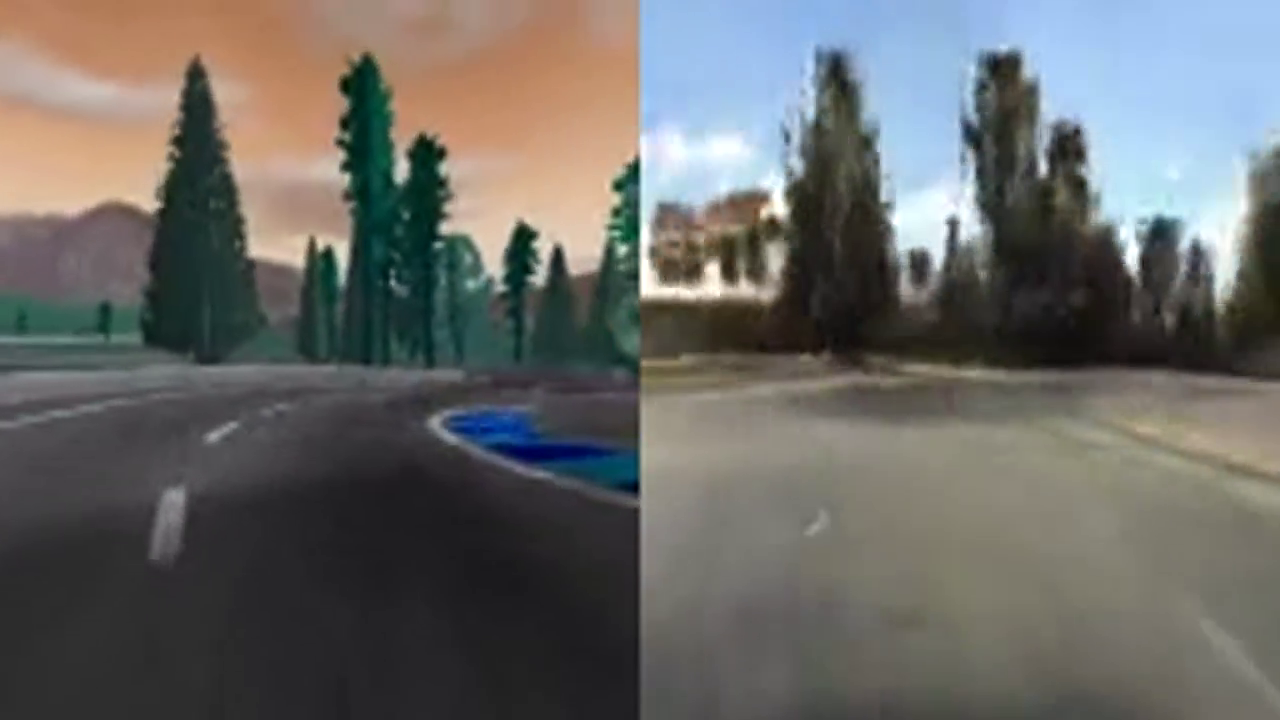}\hspace{-1.5pt}
\includegraphics[width=0.24\linewidth, height=0.125\linewidth]{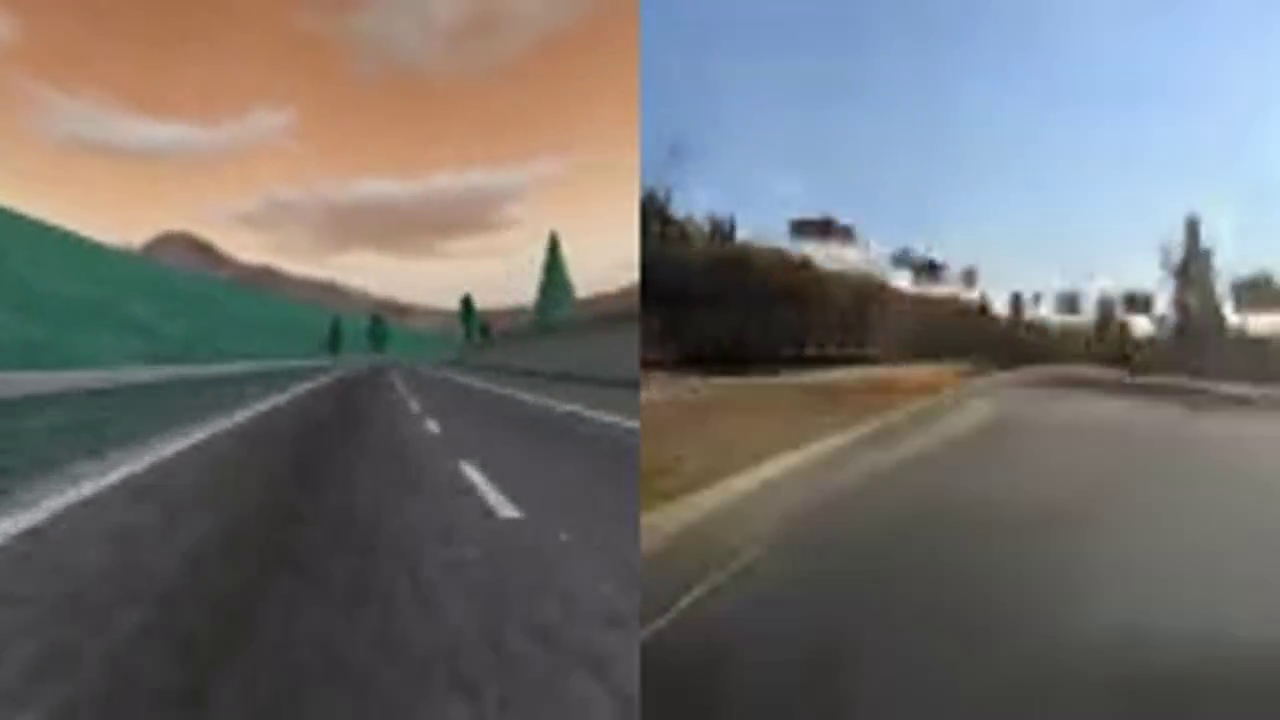}\hspace{-1.5pt}
\includegraphics[width=0.24\linewidth, height=0.125\linewidth]{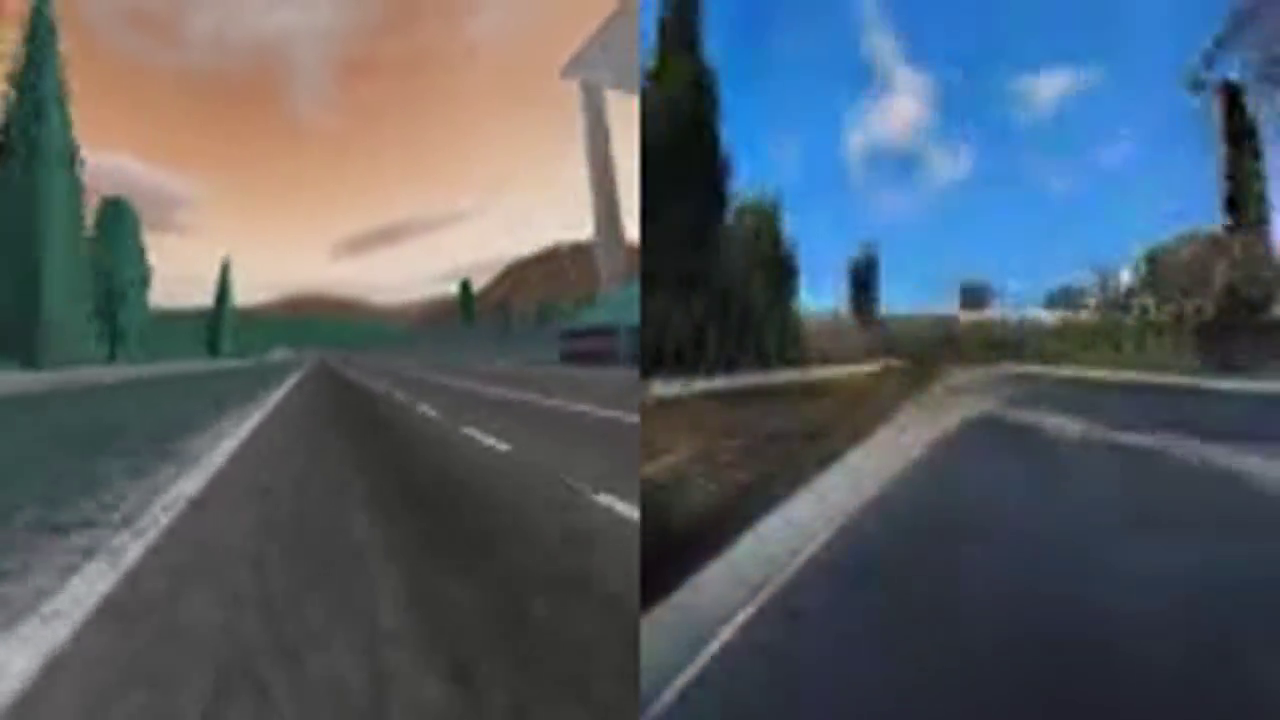}\hspace{-1.5pt}
\includegraphics[width=0.24\linewidth, height=0.125\linewidth]{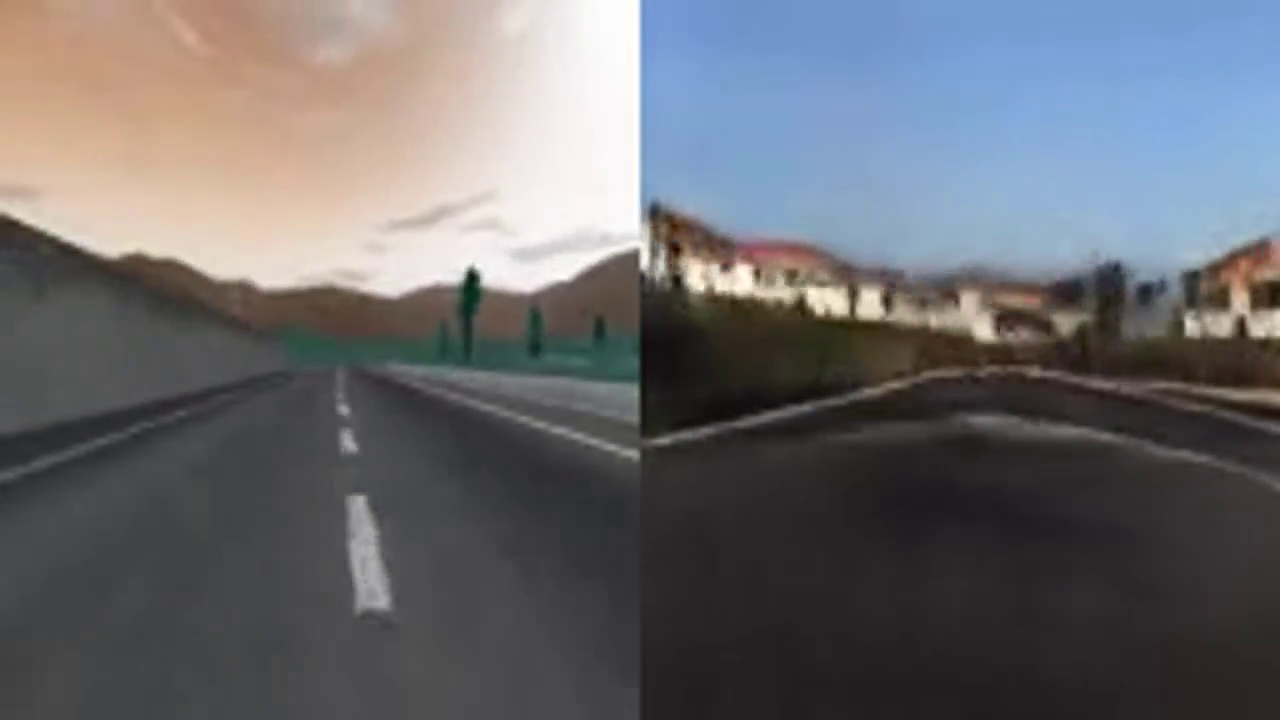}\vspace{-0.7pt}
\includegraphics[width=0.24\linewidth, height=0.125\linewidth]{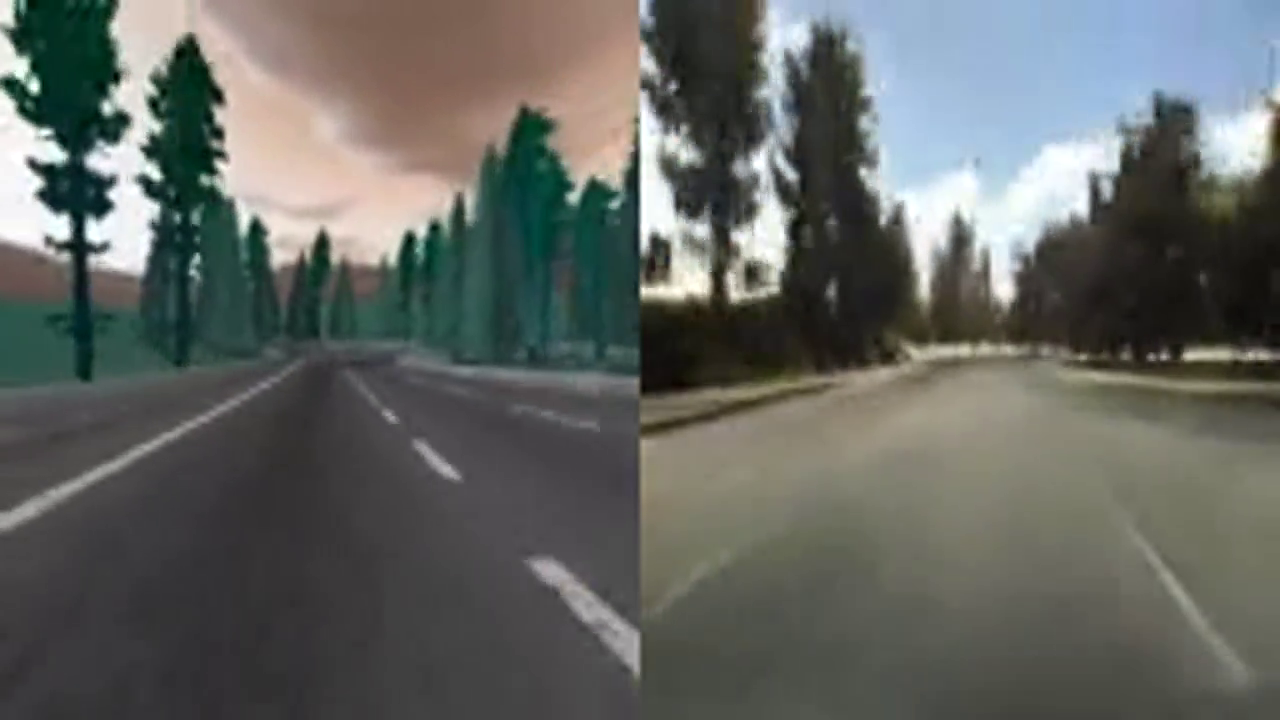}\hspace{-1.5pt}
\includegraphics[width=0.24\linewidth, height=0.125\linewidth]{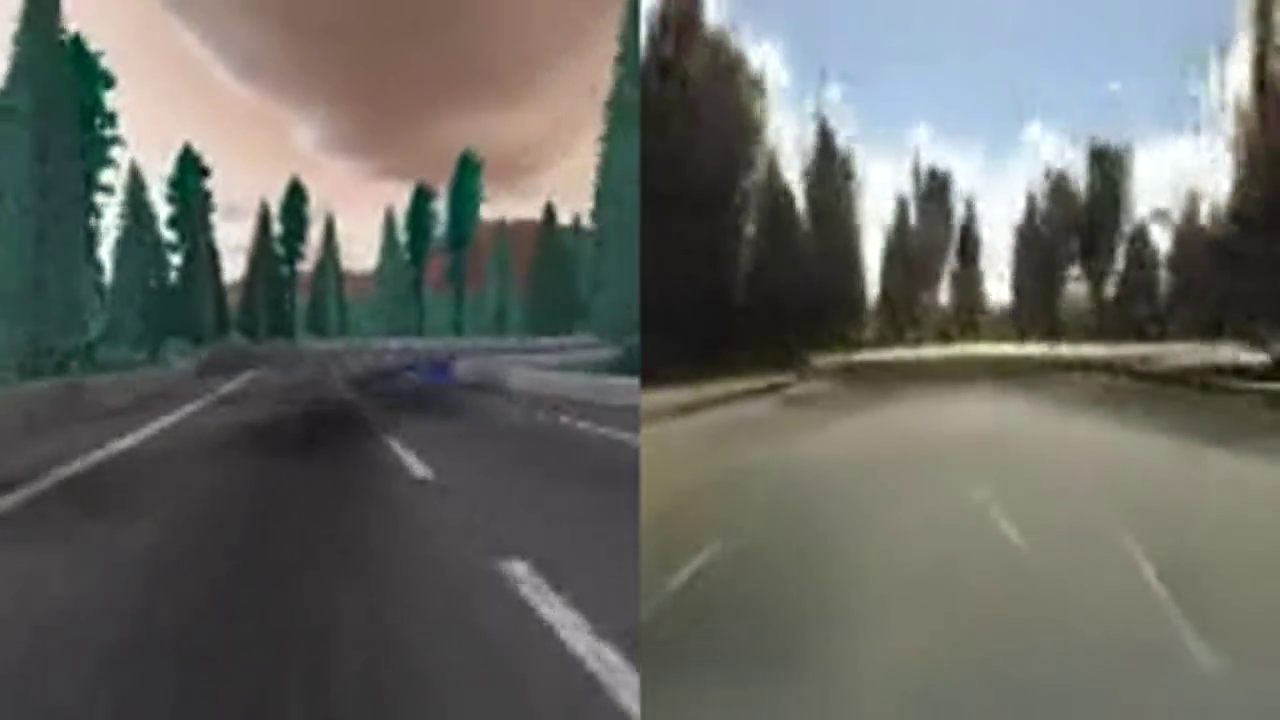}\hspace{-1.5pt}
\includegraphics[width=0.24\linewidth, height=0.125\linewidth]{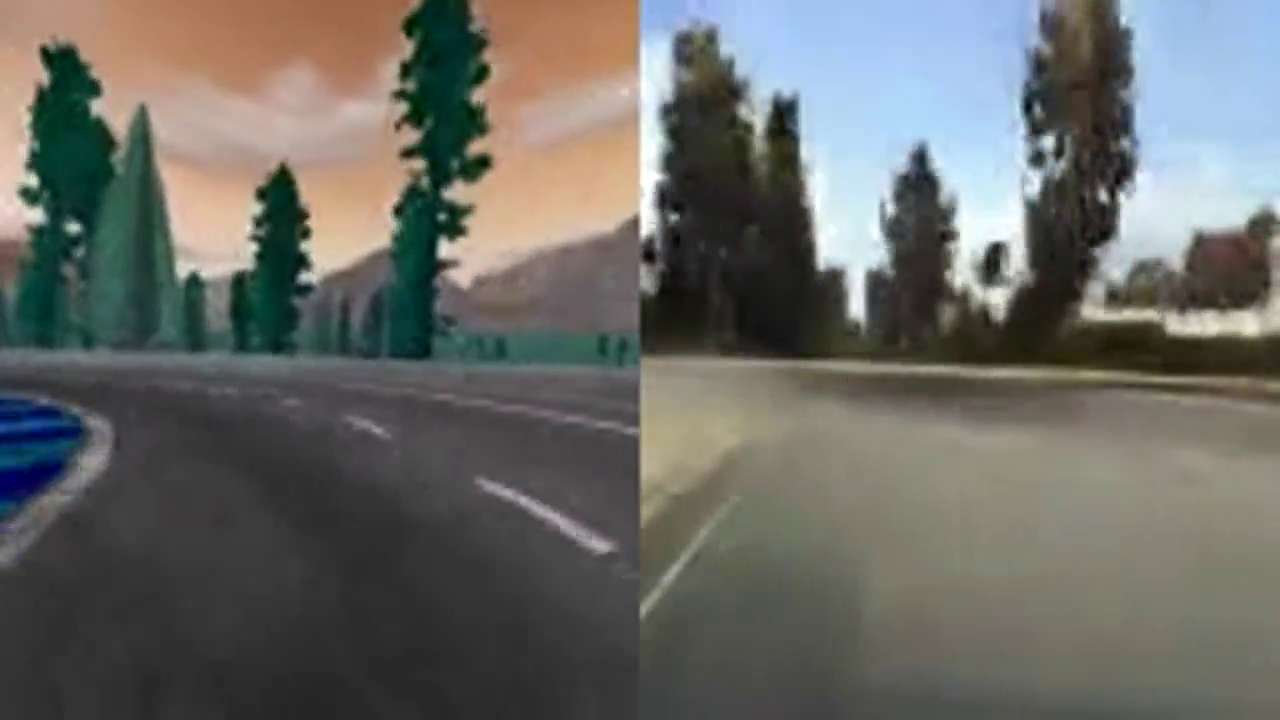}\hspace{-1.5pt}
\includegraphics[width=0.24\linewidth, height=0.125\linewidth]{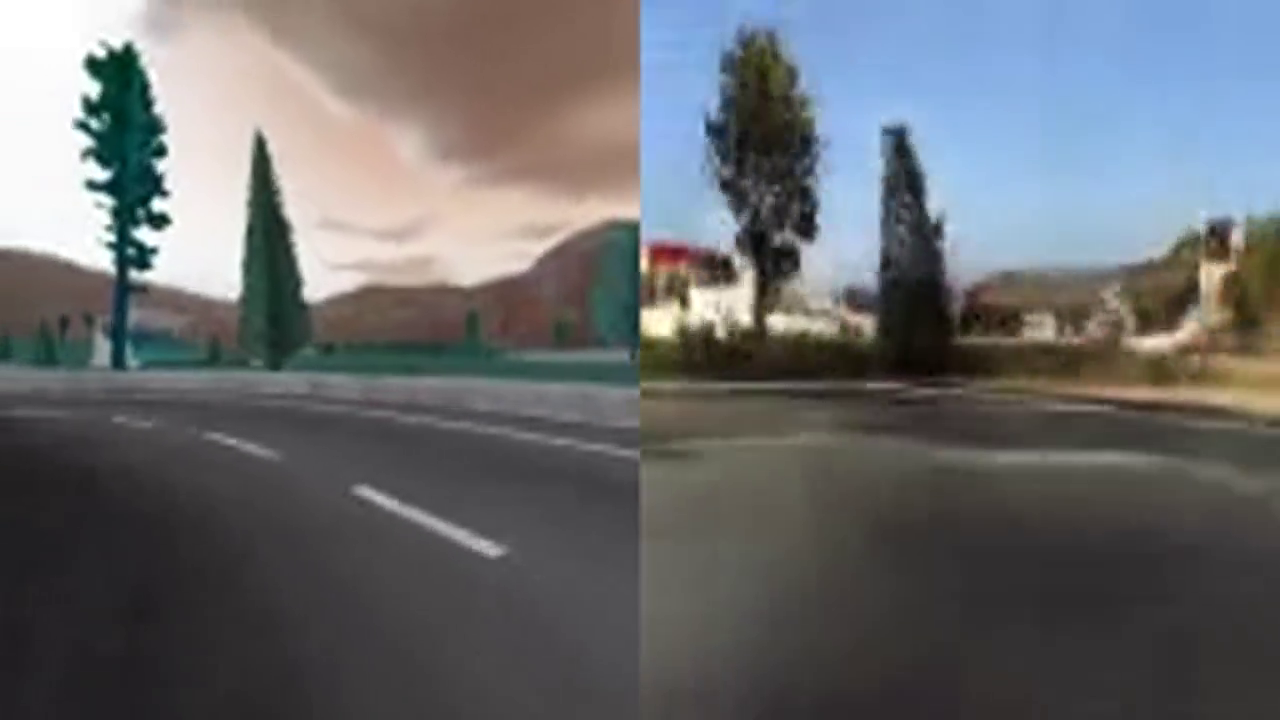}
\caption{Examples of Virtual to Real Image Translation. Odd columns
are virtual images captured from TORCS. Even columns are synthetic 
real world images corresponding to virtual images on the left.}
\label{fig3}
\end{figure*}

\begin{figure}[h]
\centering
\includegraphics[width=0.7\linewidth]{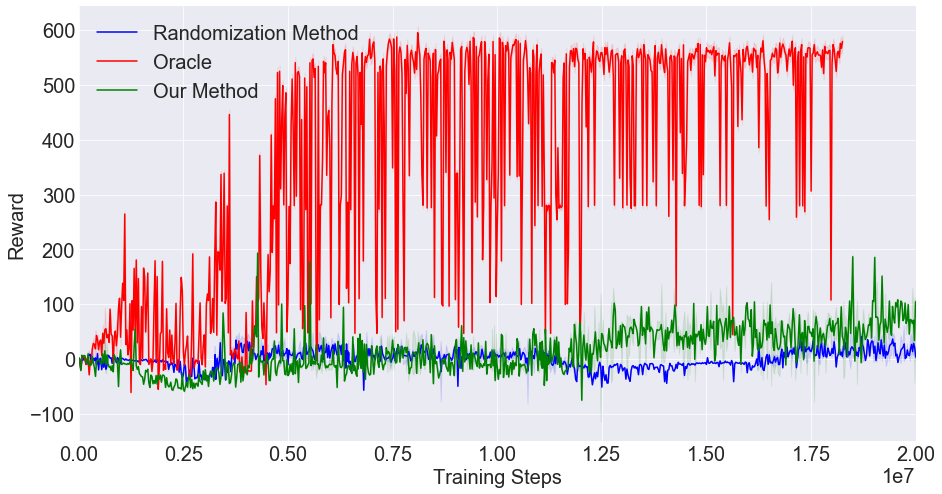}
\caption{Transfer learning between different environments. Oracle was trained in
Cgtrack2 and tested in Cgtrack2, so its performance is the best. 
Our model works better than the domain randomization RL method. 
Domain randomization method requires training in multiple virtual environments,
which imposes significant manual engineering work.
}
\label{fig4s}
\end{figure}

\section{Experiments}

We performed two sets of experiments to compare the performance 
of our method and other reinforcement learning methods as well
as supervised learning methods. The first sets of experiments
involves virtual to real reinforcement learning on real world 
driving data. The second sets of experiments involves transfer 
learning in different virtual driving environments. The virtual
simulator used in our experiments is TORCS\cite{wymann2000torcs}.

\subsection{Virtual to Real RL on Real World Driving Data}\label{exp1}
In this experiment, we trained our proposed reinforcement 
learning model with realistic translation network. We first
trained the virtual to real image translation network, and
then used the trained network to filter virtual images in 
simulator to realistic images. These realistic images were
then feed into A3C to train a driving policy. Finally,
the trained policy was tested on a real world driving data
to evaluate its steering angle prediction accuracy. 

For comparison, we also trained a supervised learning model
to predict steering angles for every test driving video frame.
The model is a deep neural network that has the same 
architecture design as the policy network in our reinforcement learning
model. The input of the network is a sequence of four consecutive
frames, the output of the network is the action probability vector,
and elements in the vector represent the probability of going
straight, turning left and turning right. The training data
for the supervised learning model is different from the testing
data that is used to evaluate model performance. In addition, another 
baseline reinforcement learning model 
(B-RL) is also trained. The only difference between B-RL and 
our method is that the virtual world images were directly 
taken by the agent as state inputs. This baseline RL is 
also tested on the same real world driving data. 

\textbf{Dataset}. The real world driving video data 
are from \cite{dataset}, which is collected in a sunny 
day with detailed steering angle annotations per frame.
There are in total around 45k images in this dataset,
of which 15k were selected for training the supervised
learning model, and another 15k were selected and held 
out for testing. To train our realistic translation 
network, we collected virtual images and their segmentations 
from the Aalborg environment in TORCS. A total of 1673 images
were collected which covers the entire driving cycle of 
Aalborg environment.

\textbf{Scene Segmentation}. We used the image semantic 
segmentation network design of \cite{badrinarayanan2015segnet} and their
trained segmentation network on the CityScape image segmentation
dataset \cite{DBLP:journals/corr/CordtsORREBFRS16} to segment 
45k real world driving images from \cite{dataset}. The network
was trained on the CityScape dataset with 11 classes and was trained
with 30000 iterations. 

\textbf{Image Translation Network Training}. 
We trained both virtual-to-parsing and parsing-to-real 
network using the collected virtual-segmentation image 
pairs and segmentation-real image pairs. The translation 
networks are  of a encoder-decoder fashion as shown 
in figure ~\ref{fig1}. In the image translation 
network, we used U-Net architecture with skip connection to 
connect two separate layers from encoder and decoder 
respectively, which have the same output 
feature map shape. The input size of the generator 
is $256\times 256$. Each convolutional layer has 
a kernel size of $4\times 4$ and striding size of $2$. 
LeakyReLU is applied after every convolutional layer 
with a slope of 0.2 and ReLU is applied after every 
deconvolutional layer. In addition, batch normalization 
layer is applied after every convolutional and deconvolutional 
layer. The final output of the encoder is connected with a 
convolutional layer which yields output of shape $3\times 256 \times 256$
followed by Tanh. We used all 1673 virtual-segmentation image pairs to 
train a virtual to segmentation network. As there are redundancies 
in the 45k real images, we selected 1762 images and their
segmentations from the 45k images 
to train a parsing-to-real image translation network. To train the 
image translation models, we used the Adam optimizer with an initial
learning rate of 0.0002, momentum of 0.5, batchsize of 16, and 200
iterations until convergence. 

\textbf{Reinforcement Training}. The RL network structure used in 
our training is similar to that of \cite{A3C} where the actor
network is a 4-layer convolutional network (shown in figure
~\ref{netarch}) with ReLU activation functions in-between. 
The network takes in 4 consecutive RGB frames as state input 
and output 9 discrete actions which corresponds to 
``go straight with acceleration'', ``go left with acceleration'',
``go right with acceleration'', ``go straight and brake'', ``go left
and brake'', ``go right and brake'', ``go straight'', ``go left'', and
``go right''. We trained the reinforcement learning agent
with 12 asynchronous threads, and with the RMSProp optimizer 
at an initial learning rate of 0.01, $\gamma=0.9$, and $\epsilon=0.1$. 

\textbf{Evaluation}. The real world driving
dataset \cite{dataset} provides the steering 
angle annotations per frame. However, the actions 
performed in the TORCS virtual environment only 
contain "going left", "going right", and "going straight"
or their combinations with "acceleration" or "brake". 
Therefore, we define a label mapping strategy 
to translate steering angle labels to action labels 
in the virtual simulator. We relate steering angle 
in $(-10,10)$ to action "going straight" (since small 
steering angle is not able to result in a distinct turning in a short time), 
steering angle less than $-10$ to action "going left" 
and steering angle more than $10$ to action 
"going right". By comparing output actions generated
from our method with ground truth, we can obtain
the accuracy of driving action prediction. 

\subsection{Transfer Learning in Virtual Driving Environments}
We further performed another sets of experiments and obtained
results of transfer learning between different virtual driving
environments. In this experiments, we trained three reinforcement
learning agents. The first agent was trained with standard
A3C in the Cg$-$track2 environment in TORCS, and evaluated
its performance frequently in the same environments. It is reasonable
to expect the performance of this agent to be the best, so we call
it "Oracle". The second agent was trained with our proposed
reinforcement learning method with realistic translation network.
However, it was trained in E-track1 environment in TORCS and then
evaluated in Cg-track2. It is necessary to note that 
the visual appearance of E-track1 is different from that of Cg-track2.
The third agent was trained with domain randomization method similar 
to that of \cite{DBLP:journals/corr/SadeghiL16}, where the agent was 
trained with 10 different virtual environments and evaluated in 
Cg-track2. For training with our methods, we obtain 15k segmented
images for both E-track1 and Cg-track2 to train virtual-to-parsing
and parsing-to-real image translation networks. The image translation
training details and reinforcement learning details are the same as that
of section~\ref{exp1}.

\section{Results}
\textbf{Image Segmentation Results}. 
We used image segmentation model trained on the cityscape
\cite{DBLP:journals/corr/CordtsORREBFRS16} dataset to segment
both virtual and real images. Examples are shown in figure
~\ref{fig2}. As shown in the figure, although the original virtual 
image and real image look quite different, their scene parsing results
are very similar. Therefore, it is reasonable to use scene parsing as the
interim to connect virtual images and real images. 

\textbf{Qualitative Result of Realistic Translation Network}.
Figure ~\ref{fig3} shows some representative results of our image 
translation network. The odd columns are virtual images in TORCS, and
the even columns are translated realistic images. The images
in the virtual environment appears to be darker than the translated
images, as the real images used to train the translation network is captured
in a sunny day. Therefore, our model succeed to synthesize realistic 
images of similar appearance with the original ground truth real images. 

\textbf{Reinforcement Training Results}.
The results for virtual to real reinforcement learning on real world
driving data are shown in table ~\ref{table:acc}.
Results show that our proposed method has a better overall
performance than the baseline method (B-RL), where the reinforcement training
agent is trained in a virtual environment without seeing any
real data. The supervised method (SV) has the best overall
performance, however, was trained with large amounts of supervised 
labeled data.

\begin{table}[h]
\centering
\caption{Action prediction accuracy for the three methods.}
\begin{tabular}{cccc}
\hline Accuracy & Ours & B-RL & SV \\
\hline Dataset in \cite{dataset} & $43.40\%$ & $28.33\%$ & $53.60\%$\\
\hline
\label{table:acc}
\end{tabular}
\end{table}

The result for transfer learning in different virtual environments is shown in 
figure~\ref{fig4s}. Obviously, standard A3C (Oracle) trained and tested 
in the same environment gets the best performance. However, our model performs better than the domain randomization method, which requires training in multiple
environments to generalize. As mentioned in \cite{DBLP:journals/corr/SadeghiL16},
domain randomization requires lots of engineering work to make it generalize.
Our model succeeds by observing translated images from E-track1 to Cg-track2, 
which means the model already gets training in an environment that looks very
similar to the test environment (Cg-track2), thus the performance is improved.

\section{Conclusion}

We proved through experiments that by using synthetic real images as training
data in reinforcement learning, the agent generalizess better
in a real environment than pure training with virtual data 
or training with domain randomization. The next step would 
be to design a better image-to-image translation network 
and a better reinforcement learning framework to surpass 
the performance of supervised learning.

Thanks to the bridge of scene parsing, virtual images can be 
translated into realistic images which maintain their scene structure. 
The learnt reinforcement learning model on realistic frames 
can be easily applied to real-world environment. 
We also notice that the translation results 
of a segmentation map are not unique. For example, segmentation 
map indicates a car, but it does not assign which color of that car 
should be. Therefore, one of our future work is to make 
parsing-to-realistic network output various possible appearances 
(e.g. color, texture). In this way, bias in reinforcement learning
training would be largely 
reduced.

We provide the first example of training a self-driving vehicle
using reinforcement learning algorithm by interacting with a
synthesized real environment with our proposed image-to-segmentation
-to-image framework. We show that by using our method for RL training,
it is possible to obtain a self driving vehicle that can be placed
in the real world.

\clearpage
\newpage
\bibliography{egbib}
\end{document}